# Predicting the post-wildfire mudflow onset using machine learning models on multi-parameter experimental data

Mahta Movasat[1], PhD.,PE, Ingrid Tomac[1,2], Associate Professor, PhD.,PE.

[1] Structural Engineering Department, University of California, San Diego, San Diego, USA

[2] Corresponding Author

itomac@ucsd.edu

The authors declare they have no financial interests.

**Abstract**

Post-wildfire mudflows are increasingly hazardous due to the prevalence of wildfires, including those on the wildland-urban interface. Upon burning, soil on the surface or immediately beneath becomes hydrophobic, a phenomenon that occurs predominantly on sand-based hillslopes. Rainwater and eroded soil blanket downhill, turning into catastrophic debris flows. Soil hydrophobicity enhances erosion, differing post-wildfire debris flows from natural mudflows in intensity, duration, and destructiveness. Thus, it is crucial to understand the timing and conditions of debris flow onset, as a result of the coupled effect of critical parameters: varying rain intensities (RI), slope gradients ($\delta$), water entry values ($\Psi_{wev}$), and grain sizes ($D_{50}$). Machine Learning (ML) techniques have become increasingly valuable in geotechnical engineering due to their ability to model complex systems without predefined assumptions. This study applies multiple ML algorithms: multiple linear regression (MLR), logistic regression (LR), support vector classifier (SVC), K-means clustering, and principal component analysis (PCA) to predict and classify outcomes from laboratory experiments that model field conditions using a rain device on various soils in sloped flumes. While MLR effectively predicted total discharge, erosion predictions were less accurate, especially for coarse sand. LR and SVC achieved good accuracy in classifying failure outcomes, supported by clustering and dimensionality reduction. Sensitivity analysis revealed that fine sand is highly susceptible to erosion, particularly under low-intensity, long-duration rainfall. Results also show that the first 10 minutes of high-intensity rain are most critical for discharge and failure. These findings highlight the potential of ML for post-wildfire hazard assessment and emergency response planning.

1. Introduction

Post-wildfire hydrophobicity has become a growing concern in recent decades, as the frequency, intensity, and duration of wildfires have increased due to climate change and land use changes (Westerling et al., 2006). One of the major hazards associated with post-fire hydrophobic soils is the occurrence of debris flows. These flows are dangerous because they can be triggered within minutes of rainfall initiation and require significantly less rainfall to occur in burned areas than in unburned areas (Cannon et al., 2008; Kean et al., 2011). The resulting debris flows pose substantial threats to life, infrastructure, and the environment. Water repellency (hydrophobicity) in soil refers to the reduced affinity of the soil surface to absorb water. This phenomenon occurs when the surface free energy of soil grains is lower than the surface tension of water, leading water to bead on the soil surface rather than immediately infiltrate. Conversely, when the grain surface free energy exceeds the water's surface tension, water spreads and is absorbed by the soil (Doer et al., 2000; Leelamanie et al., 2008). Hydrophobic soils can lead to the formation of preferential surficial flow paths, increased splash detachment, surface runoff, and, in certain cases, debris flows—especially in post-fire landscapes (Debano, 1981, 2000; Martin et al., 2001; Nyman et al, 2011; Ritsema & Dekker, 2012). The complexity and uncertainty of post-wildfire soil behavior, particularly its hydrological and mechanical responses, make it a suitable domain for advanced modeling approaches. In this context, the rapid development of computer technology and the growing availability of large, high-dimensional datasets have catalyzed the application of machine learning (ML) techniques across a broad range of scientific and engineering fields. In geotechnical engineering, where many influencing parameters are interdependent, ML provides a robust, data-driven alternative to traditional deterministic models (Goh & Zhang et al, 2014; Wang et al., 2020; Zang et al., 2021). Unlike conventional models that rely on predefined physical assumptions and relationships, ML algorithms can directly identify patterns, correlations, and underlying trends from data. ML improves models' predictive capabilities through training and generalizes to unseen scenarios, offering considerable advantages in uncertainty-prone environments (El Naqa et al., 2015; Goh et al., 2018; Zhang et al., 2019). This adaptability has led to successful applications of ML in various geotechnical problems, including soil classification (Bhattacharya & Solomatine, 2006), cone penetration test (CPT) interpretation (Goh, 1995), slope stability evaluation, debris flow likelihood prediction (Erzin & Cetin, 2013; Staley et al., 2016; Di Napoli et al., 2020; Tiend Bui et al., 2019, Kern et al., 2017; Nikolopoulos et al., 2018; Qi & Tang, 2018; Zhang et al., 2019), and estimation of geotechnical parameters such as SPT-N value relationships and compression index (Puri et al., 2018; 2020). Logistic regression models are frequently used to predict debris-flow likelihood across various studies and regions (Cui & Cheng, 2019; Kern et al., 2017; Nikolopoulos et al., 2018; Staley et al., 2017; Addison & Oommen, 2020; Cannon et al., 2010; Staley et al., 2016). Some of the variables in these studies include soil properties, such as the percentage of clay and liquid limit, average storm intensity, and geospatial properties of the region that characterize the basin morphology, including the gradient of the burnt area with different severities and ruggedness. Algorithms such as artificial neural networks (ANNs), support vector machines (SVMs), and random forests (RFs) have proven particularly effective in tackling nonlinear and complex geotechnical datasets (James et al., 2013).

In this study, machine learning algorithms—logistic regression (LR), multiple regression (MR), and support vector machines (SVM)—are used to model and predict geotechnical outcomes influenced by soil hydrophobicity and post-wildfire conditions. These methods are chosen for their strong predictive performance, wide adoption in industry, and proven ability to handle both classification and regression problems in complex, nonlinear environments (Qi & Tang, 2018; James et al., 2013). While previous studies have applied ML to post-wildfire debris flow prediction, they typically rely on field data, where factors such as soil properties, burn severity, and rainfall intensity are difficult to control and highly variable. This study addresses that gap by using a more controlled experimental environment in which soil hydrophobicity, grain size, slope, and rainfall intensity are systematically varied in controlled laboratory conditions.

## 2. Materials and Methods

### 2.1. Experimental Data

The experimental framework was designed to investigate how key physical parameters influence hydrophobic soil behavior and post-wildfire slope stability under simulated rainfall conditions. Experimental flumes mimic field conditions, where artificial hydrophobic sand layers are placed on top and below the soil surfaces, are inclined at different angles at times, and subjected to a range of rain intensities and durations (Fig.1). Specifically, this study uses 36 experimental flume tests (Table 1) (Movasat, 2022; Movasat & Tomac, 2024) at two site-specific tipical soil configurations to investigate the effects of a hydrophobic layer's position within the top 2 cm of soil:

1. H-Top layout: A 2 cm hydrophobic layer is placed directly on the soil surface.
2. H-Sub layout: A 1 cm hydrophobic layer is embedded beneath a 1 cm hydrophilic layer.

Both configurations use a 13 cm hydrophilic base, and the selected layer thicknesses are based on literature, field relevance, and experimental feasibility. The lab-scale hydrophobic layer thickness is similar and has the same order of magnitude to the soil layer thickness in field-scale post-wildfire hillslopes. Post-wildfire debris flows erode burned scars, where fire-induced hydrophobic soil layers have an average depth of 0.5-5 cm (Parsons et al. 2010). Although the experimental setup size is much smaller than the scale of field events, the goal of the research was to better understand the origins and mechanisms of slope instability and debris-flow initiation. Since most debris flows originate from different erosional rill environments, smaller experimental scales refer only to the surface area. They investigate, at close range, the origins of erosional processes on sandy surfaces with original full-scale slope angles, sand particle sizes, and raindrop sizes. The 1 cm hydrophobic thickness was sufficient to observe erosion and ensure consistency across tests. Equal thickness of hydrophobic and hydrophilic materials also removes the effect of thickness variation on readings from water content sensors. The sensor layout is shown in Fig. 1 with small rectangle shapes numbered from 1-4.

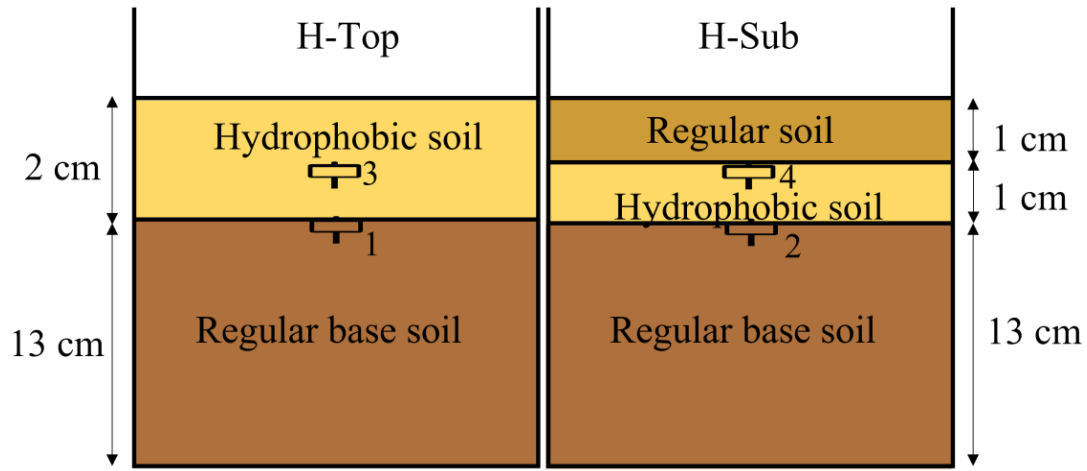

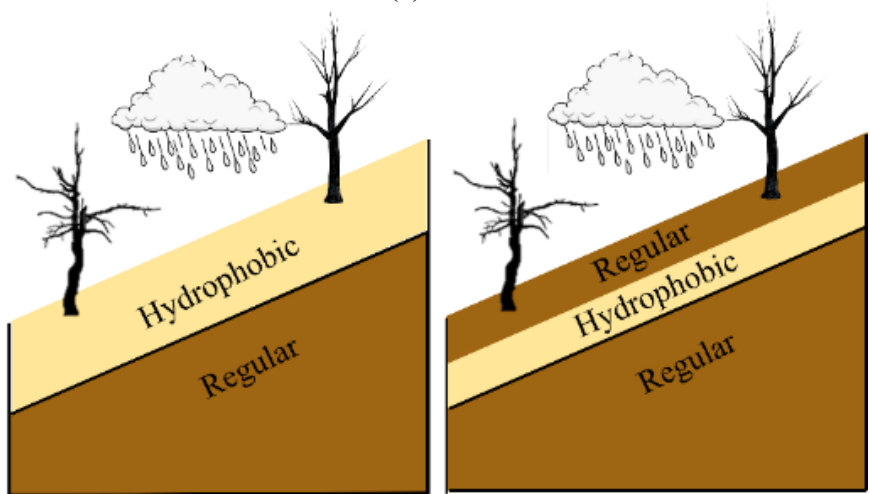

Fig. 1 Layout of the two flumes, a) dimensions of the cross section across the flume, and b) cross section along the flume (non-scaled schematics).. Left flume represents the H-Top layout, and the right flume represents the H-Sub layout The water content sensors 1-4 embedded in the soil are shown with rectangular boxes in the sketches.

Table 1- Input parameters for simulated rainfall experiments, each slope gradient uses H-Top and H-Sub layer configurations.

| Soil | Friction angle, $\varphi'$ (°) | Rain Intensity, (mm/hr) | Slope gradient, $\delta$ (°) |
|---|---|---|---|
| Fine ($D_{50}$=0.2 mm) | 30 | 18 | 20 |
| | | | 30 |
| | | 70 | 20 |
| | | | 30 |
| | | 120 | 20 |
| | | | 30 |
| Medium ($D_{50}$=0.4 mm) | 32 | 18 | 20 |
| | | | 30 |
| | | 70 | 20 |
| | | | 30 |
| | | 120 | 20 |
| | | | 30 |
| Coarse ($D_{50}$=0.65 mm) | 34 | 18 | 20 |
| | | | 30 |
| | | 70 | 20 |
| | | | 30 |
| | | 120 | 20 |
| | | | 30 |

## 2.2 Machine Learning Analysis

Four main input features—median grain size ($D_{50}$), water entry value ($\Psi_{wev}$), slope gradient ($\delta$), and rain intensity (RI)—were derived from controlled rainfall experiments. Two experimental layouts, H-Top and H-Sub, were analyzed separately to represent different configurations of hydrophobic and regular soil layers. For the first group, the H-Top layout, we attempted to predict the total water discharge and erosion rates using a multiple linear regression model. For the second group, the H-Sub layout, the goal is to classify outcomes and determine whether the model experiences an infinite failure. First, a clustering method combined with principal component analysis is used to divide the H-Sub data into two clusters, and a support vector classifier is then used to determine the maximal-margin hyperplane separating the two classes. A logistic regression method is also used to classify the data based on the outcomes and to determine the probability and likelihood of failure given the parameters. For both H-Top and H-Sub layout analysis, four influencing features are median grain size ($D_{50}$) in mm, water entry value ($\Psi_{wev}$) in kPa, slope gradient ($\delta$) in degrees, and rain intensity ($RI$) in mm/hr. An outline is shown in Fig. 2 to summarize the technical route and the applied methods.

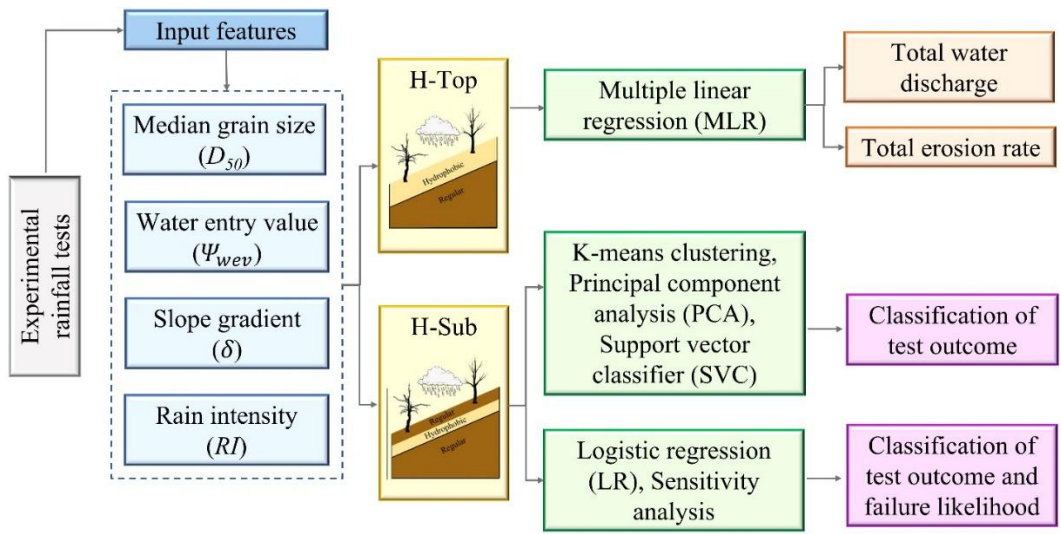

Fig. 2 Outline of the technical route for this study

## 3. Results

### 3.1. Multiple linear regression (H-Top layout)

In the first part of the study, a multiple linear regression (MLR) model is used to predict the total amount of water discharge and erosion rate. The data include the 18 experimental results from H-Top layout tests, in which a hydrophobic layer of soil rests on the flume's top surface. The data consists of two sets: training data for training the MLR model and testing data for testing the trained model. The ratio of testing to training data has been set to 35% to ensure an accurate model, given the limited data from experimental results.

Different features such as median grain size (mm) ($D_{50}$), the effective size of soil (mm) ($D_{10}$), coefficients of curvature and uniformity of sand ($C_c$, $C_u$), contact angle $\theta$ (°), friction angle $\varphi$ (°), water entry value (kPa) ($\Psi_{wev}$), slope gradient $\delta$(°), and rain intensity (mm/hr) ($RI$) are introduced to the model, and a correlation matrix is obtained to visualize the correlation of the input features with output results (Fig. 3). The test outputs such as total cumulative erosion rate (gr/m$^2$) and total cumulative water discharge (lit/m$^2$) are shown with *TE* and *TD* symbols in the correlation matrix. In addition, the erosion rate in the first, second, ..., sixth 10-minute intervals is shown as 1E, 2E,..., 6E (gr/m$^2$·min), respectively, and water discharges are shown as *1D, 2D, ..., 6D* (lit/m$^2$·min ). The six 10-minute intervals are chosen based on observations on smaller sand pods subjected to rain from nozzles during the initial stages of the experimental design. The 10-minute interval is chosen to be shorter than the standard rain gauge measurement interval for real rain data ("15 minutes"). The 60-minute time point was chosen based on experimental observations, as no changes were observed beyond that time.

Although shorter intervals for collecting overflow water and eroded sediment may yield greater accuracy near the peak of erosion, we chose a consistent pattern to cover the full range of erosion rates. We also observed that the peak erosion time was well captured by collecting water and eroded sand at 10-minute intervals. Larger intervals would prevent any insights and would completely miss the peak erosion time. Furthermore, in the initial and later stages of the experiment, shorter intervals were not chosen due to very low erosion rates and smaller sediment amounts that were difficult to collect. Complete graphs for all experiments showing erosion rates and overflowed water versus time, including peak values, can be found in Movasat and Tomac (2024). The correlation matrix shows that the contact angle ($\theta$) and water entry value ($\Psi_{wev}$) show a similar correlation with different outputs, and therefore the contact angle is removed from the features. The correlation between $D_{50}$ and $D_{10}$ with the discharge and erosion rates is very similar. The $D_{10}$ values are 1.3~1.4 times smaller than the $D_{50}$ values, and this ratio is almost the same for all sands. Since the proportionality of $D_{50}$ and $D_{10}$ values is constant for all sands, the correlations of results are not affected. The coefficient of curvature ($C_c$) has a similar correlation to $D_{10}$ and $D_{50}$. The correlation of $C_c$, $D_{10}$, and $D_{50}$ is stronger for erosion rate than for discharge values.

It is shown that while $\Psi_{wev}$, $\delta$, and *RI* are positively correlated with discharge and erosion, $D_{50}$ shows a negative correlation. The correlations between the total discharge amount (*TD*) and the input features show that *RI* has the highest absolute correlation value (0.77), followed by $\delta$ (0.44). $D_{50}$ and $\Psi_{wev}$ have similar absolute correlation values (0.28) with *TD*. The absolute correlation value of RI (0.5) decreases for total erosion rate (*TE*), while the absolute correlation value of $D_{50}$ (0.59) and $\Psi_{wev}$ (0.52) increases, and the correlation of $\delta$ (0.4) stays almost unchanged. The correlation between rain intensity (RI) and erosion and discharge values indicates that RI's correlation is stronger at the beginning of the tests and decreases over time. The slope angle ($\delta$) shows an almost constant correlation with erosion rate over time, while becoming stronger in correlation with discharge as time passes. The correlation matrix is beneficial in depicting the overall influence of different features on the outputs.

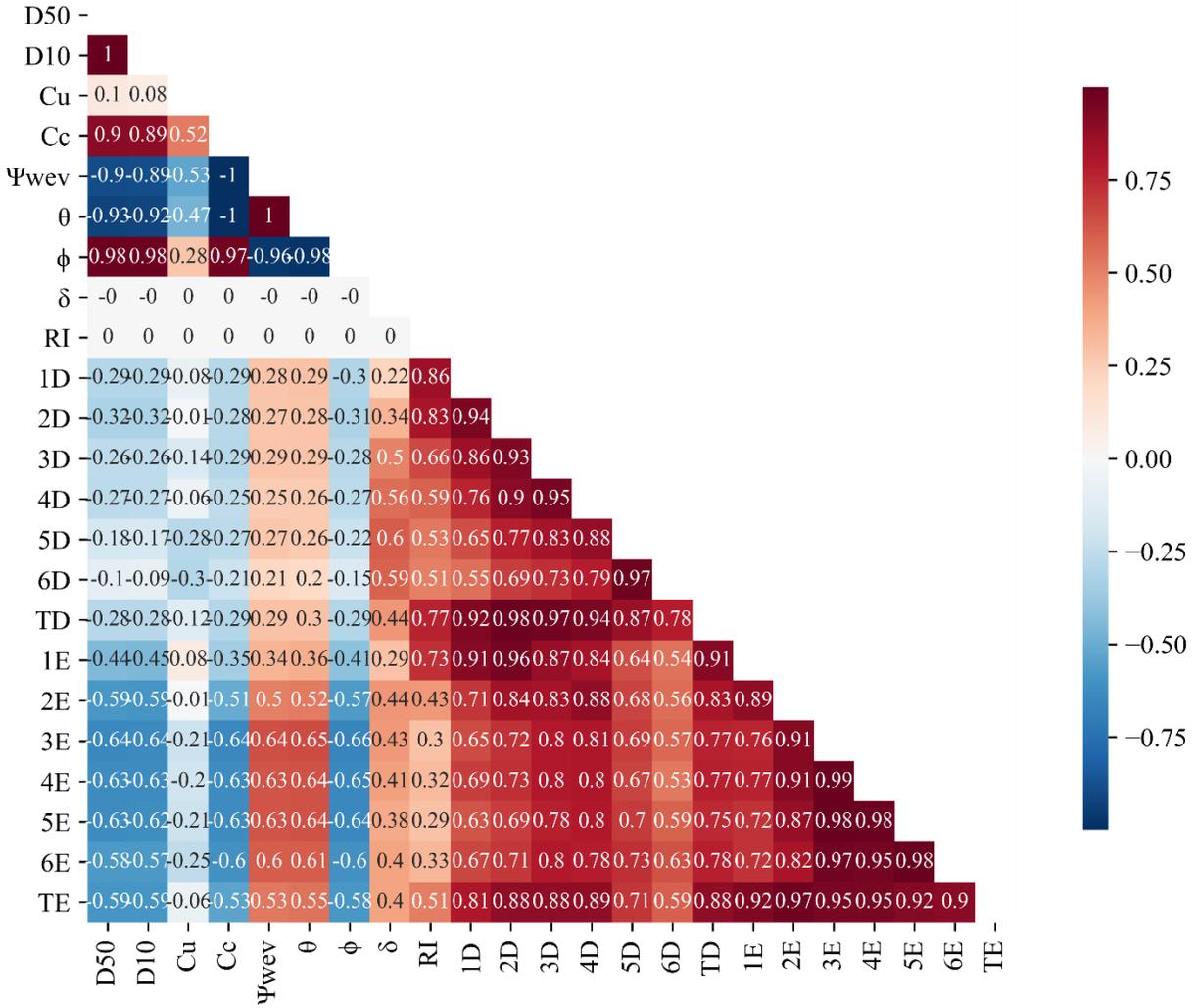

Fig. 3 Correlation matrix for H-Top results

A multiple linear regression approach predicts total discharge and erosion rates. The initial six features are reduced to four final features, $D_{50}$, $\Psi_{wev}$, $\delta$, and $RI$, to train the data since the elimination of contact angle ($\theta$), and friction angle ($\varphi$) did not affect the results of the predictive capability of the MLP models. Theoretically, the water discharge and erosion rate should be close to zero if there is no rainfall and the intensity ($RI$) is set to zero. For that reason, all other features, $D_{50}$, $\Psi_{wev}$, and $\delta$, are multiplied by $RI$, so that in the absence of rain ($RI$=0), the output of discharge and erosion rate is close to zero. Some researchers have recently applied the same approach to post-fire debris-flow likelihood studies (Nikolopoulos et al., 2018; Staley et al., 2017; Addison & Oommen, 2020). The following equation (Eq. 1) presents the multiple linear regression used in this study. In this equation, $X_1$ is the first feature and denotes $D_{50}.RI$, $X_2$ denotes $\Psi_{wev}.RI$, and $X_3$ is $\delta.RI$. The rainfall intensity is based on one-hour accumulation:

$$y(X) = \lambda_0 + \lambda_1 X_1 + \lambda_2 X_2 + \lambda_3 X_3 \qquad (1)$$

Table 2 shows the variables and their optimal coefficients in the trained model. The models are trained based on 65% of the total data. 35% of the remaining data is used for testing the model. To evaluate the model's performance, the mean squared errors and $R^2$ scores for each model, using equations (Eqs. 2 and 3) for training and testing data, are shown in Table 2. The $R^2$ and MSE were chosen deliberately because they provide complementary perspectives on model performance. R2 quantifies the proportion of variance explained by the model, while MSE emphasizes larger prediction errors through its squared term, which is important for assessing the model's reliability in potentially higher-error applications, such as erosion data. MAE can offer a more interpretable average error magnitude and is less sensitive to outliers. Given the skewness in the erosion data, we believe MSE's sensitivity to larger errors is more appropriate for capturing the model's performance on extreme values.

$$R^2 = 1 - \frac{\sum_{i=1}^{n}\left(y_{i(Actual)} - \hat{y}_{i(Predicted)}\right)^2}{\sum_{i=1}^{n}\left(y_{i(Actual)} - \bar{y}_{i(mean)}\right)^2} \qquad (2)$$

$$MSE = \frac{1}{n}\sum_{i=1}^{n}\left(y_{i(Actual)} - \hat{y}_{i(Predicted)}\right)^2 \quad (3)$$

Table 2. Summary of coefficients and statistical evaluation parameters for testing and training data of the multiple linear regression models for total discharge (TD) and total erosion (TE).

| Features | Coefficients | Total Discharge (*TD*) (Duration=60 min) | Total Erosion (*TE*) (Duration=60 min) |
|---|---|---|---|
| | $\lambda_0$ | 11.3 | -15.2 |
| $X_1$ ($D_{50}.RI$) | $\lambda_1$ | -0.46 | -90.7 |
| $X_2$ ($\Psi_{wev}.RI$) | $\lambda_2$ | 0.025 | -5.2 |
| $X_3$ ($\delta.RI$) | $\lambda_3$ | 0.025 | 2.9 |
| Evaluation | | $R^2$ train: 0.96  $R^2$ test: 0.72  MSE train: 14.5  MSE test: 75.3 | $R^2$ train: 0.87  $R^2$ test: 0.45  MSE train: 655911  MSE test: 1070667 |

Figs. 4 and 5 compare the training and testing data collected from experimental data with predicted results from MLR models. The model evaluation is also shown in Table 2 for both testing and training predictions. The MLR model captures total discharge (TD) with $R^2$ values of 0.96 and 0.72 for training and testing data, respectively. This indicates lower variation in the output-dependent attribute, as predicted by the independent input variables. On the other hand, the MLR model for total erosion (*TE*) has an $R^2$ of 0.87 on the training data, but decreases to 0.45 on the test data. This means that approximately 45% of the observed variation can be explained by the model's inputs and suggests that the model is likely overfitted. A larger testing set could lead to lower $R^2$ values while the total data set remains relatively small Although we concluded that the model is likely overfitted based on the $R^2$ test data for total erosion, this reference points to the applicability of the proposed ML approach, not to the reality of the physical experiments, which have shown a clearer dependence of erosion on grain size (Movasat and Tomac, 2024).

The general purpose of MLR is to learn more about the relationship between the independent ($D_{50}$, $\Psi_{\text{wev}}$, $\delta$, $RI$) variables and a dependent or criterion variable, which is the total discharge (*TD*) and total erosion (*TE*) in this case. While the total discharge (*TD*) model performed well, the total erosion (TD) model did not. It is observed in Fig. 6 (a) that the predicted total discharge amounts are closer to their actual values both in training and testing data points. However, the actual and predicted total erosion data are more scattered around the identity line (Fig. 6(b)).

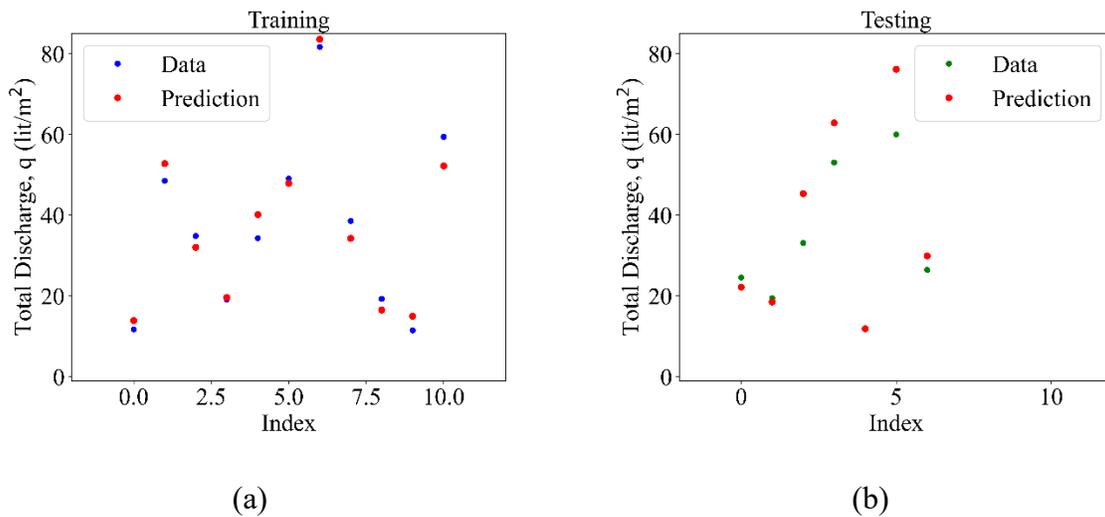

(a) (b)

Fig.4. Training data (blue symbols) (a) and testing data (green symbols) (b), and predictions (red symbols) of MLR model for total discharge

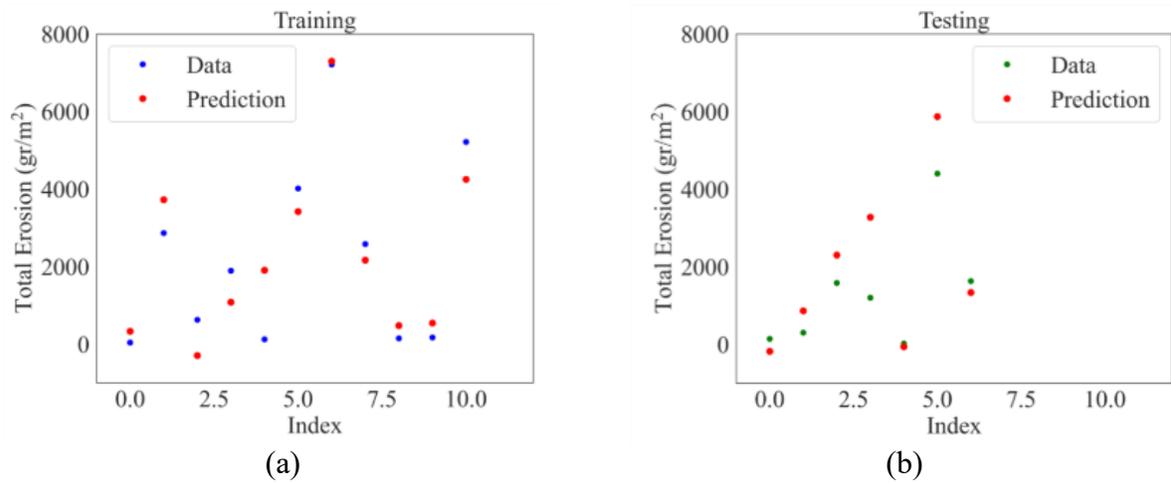

(a)             (b)

Fig.5 Training data (blue symbols) (a) and testing data (green symbols) (b) and predictions (red symbols) of MLR model for total erosion

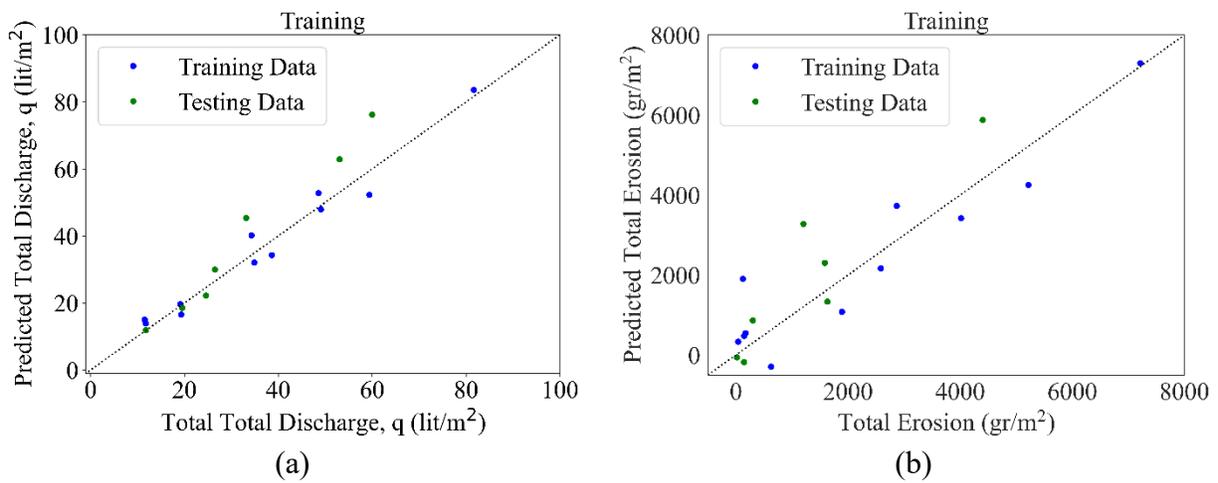

(a)             (b)

Fig.6 Performance of MLR models in predicting the training and testing data for (a) total discharge and (b) total erosion

The lower accuracy of the MLR model for erosion may be related to the low erosion values in coarse sand. Values close to zero affect the overall regression fit and can even lead to negative predictions for tests with smaller variable values.

## 3.2. K-means cluster and Principal Component Analysis (PCA) (H-Sub layout)

The K-means clustering method is straightforward for separating a data set into $k$ clusters. In this study, the K-means clustering method is used to cluster the H-Sub layout data into two clusters: those that experienced infinite failure and those that did not. Eighteen data sets are available from the experiments. The input variables of the model are median grain size ($D_{50}$), water entry value ($\Psi_{wev}$), friction angle ($\varphi$), contact angle ($\theta$), slope gradient ($\delta$), and rain intensity ($RI$).

Prior to clustering, a principal component analysis (PCA) is performed to reduce the number of input variables while preserving the variability of the original set.

It is worth noting that the variables are standardized separately by subtracting the mean and dividing by the standard deviation, thereby shifting the distributions to have a mean of zero and a standard deviation of one. Fig. 7 shows the cumulative explained variance as a function of the number of components. The analysis shows that the first two principal components together can explain 75% of the variance in the data. The last four principal components explain only 10% of the variance. We selected two principal components due to the good variance coverage and easier visualization of the two components in 2D plots. One of the main objectives of the machine learning analysis for this study is to enhance visualization and synthesize results; thus, having two principal components helps us achieve this goal.

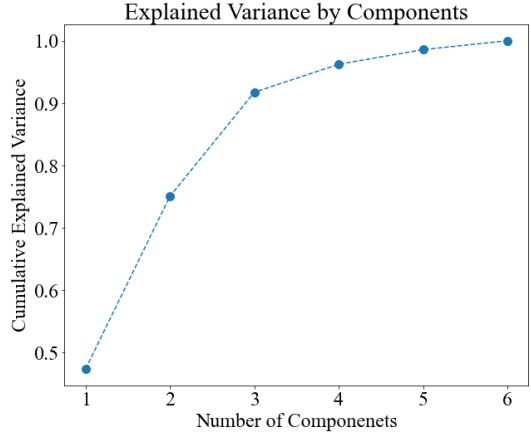

Fig.7. Cumulative explained variance versus the number of components.

After the PCA analysis and selection of the first and second principal components ($Z_1$ and $Z_2$), the K-means clustering is performed for the dataset. For transparency, the loadings of the first two principal components are reported here:

PC1=−0.402D50+0.39Ψwev+ 0.235δ +0.303RI+0.495TD+0.54TE.

PC2= −0.54D50+0.544Ψwev- 0.325δ -0.405RI-0.363TD-0.094TE.

It should be noted that these loadings are derived from a controlled laboratory dataset with a limited number of experimental configurations. As such, the numerical values of the loadings are inherently dependent on the specific variables and their ranges in this study. Consequently, the loadings should not be interpreted as universally reproducible constants, but rather as indicators of which variables are most influential and the general direction of their effect in this dataset.

Fig. 8 (a) shows the two clusters with brown ("*No failure*" = 0) and blue ("*Infinite failure*" = 1). The experimental results are also shown in Fig.8(b) with two principal components and with the same color codes. The results show that the K-means clustering method has successfully separated the two classes, and only two data points are mistakenly clustered as class 0 ("No failure"; marked with a red circle), whereas they failed in the infinite-failure experiments.

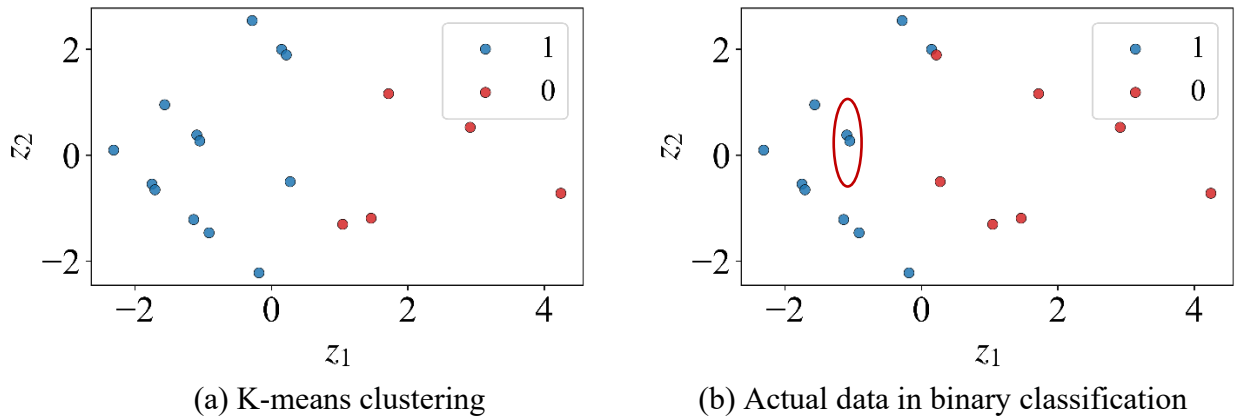

(a) K-means clustering       (b) Actual data in binary classification

Fig.8. Binary classification of experimental outcomes as ("*No failure*" = 0) and ("*Infinite failure*" = 1) with (a) K-means clustering method. (b) Binary classification of experimental outcomes as ("*No failure*"= 0) and ("*Infinite failure*"= 1)

### 3.3. Support vector classifier (SVC)

After reducing the feature dimension with PCA and clustering the data, this time the support vector classifier (SVC) is used to learn a decision boundary separating the two output classes. The two classes are labeled 1 for cases with infinite failure and 0 for cases with no failure. A linear kernel is used for the decision boundary between the two classes. The regularization parameter ($C$) is used for the penalty parameter of the error term. In other words, it controls the trade-off between accurately classifying training points and creating a smooth decision boundary, suggesting that the model chooses which data points to use as support vectors.

Fig. 9 (a-c) shows the effect of the parameter $C$ on the width of the margin and the number of support vectors. Support vectors are the data points that support the decision boundary. It is shown that when $C = 10$ (Fig. 9(a)), the margin width is smaller, and there are 5 support vectors, all from class 0 (brown data points). As $C$ becomes smaller, the margin width increases, and the number of support vectors grows. Consequently, the model selects more data points as support vectors, leading to higher variance, lower bias, and potentially overfitting. In the case of large $C$, we also encounter underfitting, leading to high bias and low variance. For this case, $C=1$ is the most appropriate selection.

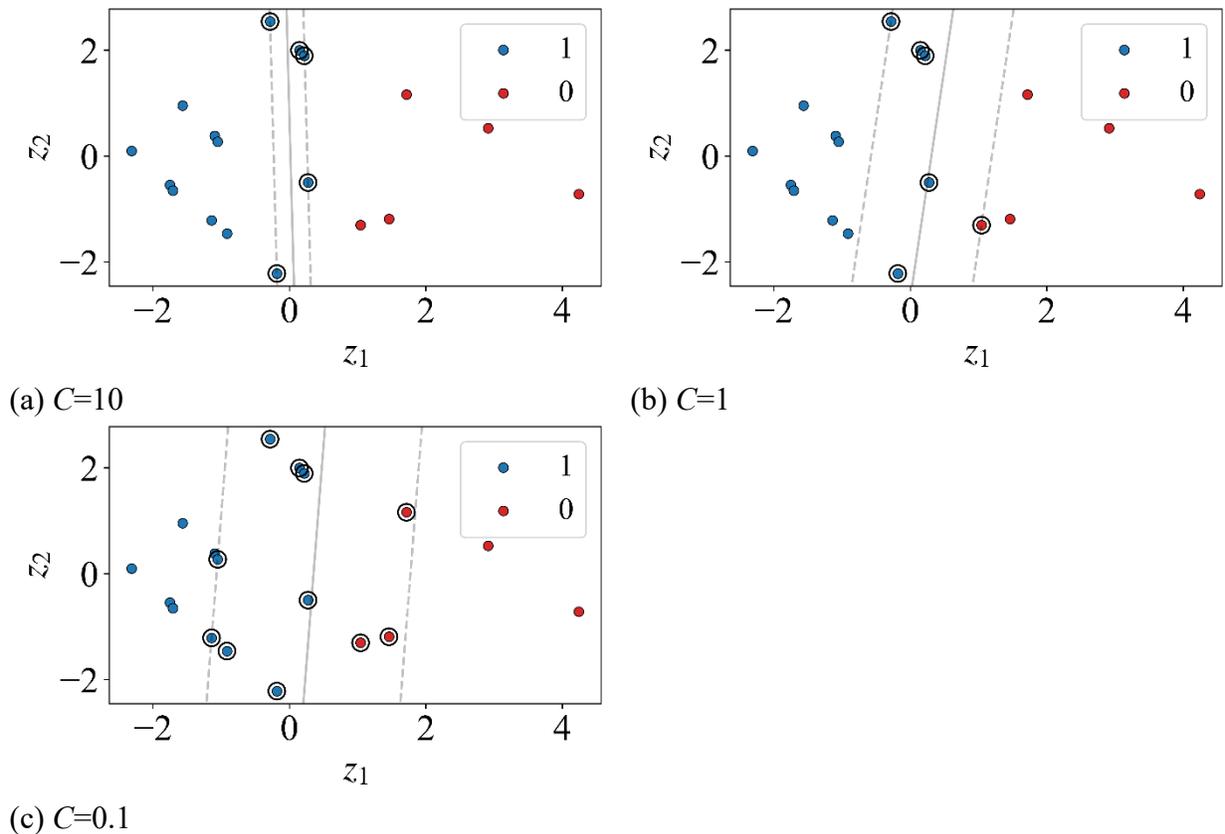

(a) $C=10$
(b) $C=1$
(c) $C=0.1$

Fig..9 Binary classification of experimental outcomes as ("*No failure*" = 0) and ("*Infinite failure*" = 1) with SVC classifier with linear kernel.

In this study, the experimental results are classified into two classes: the first class includes tests that did not fail ("No failure: 0"), and the second class includes tests in which infinite failure occurred ("*Infinite failure: 1*"). *TP*s correspond to the number of failed tests that the model correctly predicts. *FP*s are the number of tests that did not fail in the experiment but were predicted as failed by the model. *FN*s correspond to tests that failed in the experiment, but the model missed the failure and predicted it as stable. *TN*s represent the number of correctly predicted "No failure" outcomes by the model, which did not fail in the experiment.

Using the four above-mentioned classes, the performance of models is evaluated with precision, accuracy, and threat score (*TS*) (Eq (4-6)). Precision measures how well the model can detect a correctly labeled TRUE class (*TP*) from all those predictions that were predicted TRUE (*TP+FP*). Accuracy measures how well the model can detect a true labeled TRUE (*TP*) and a true label FALSE (*TN*) (Zhang et al., 2019). The threat score (*TS*) indicates the model's overall performance by reducing its value for each incorrect prediction (*FP* or *FN*):

$$\text{Precision} = \frac{TP}{TP + FP} \tag{4}$$

$$\text{Accuracy} = \frac{TP + TN}{TP + FP + TN + FN} \tag{5}$$

$$\text{Threat score } (TS) = \frac{TP}{TP + FP + FN} \tag{6}$$

Fig. 10 shows the SVC model with a 35% training/test split and *C*=1. The choice of (*C* = 1) was guided by Fig. 9, which shows how the margin width and number of support vectors vary with (*C*). This value provides a balanced trade-off between underfitting and overfitting. The selection is further supported by training/test splits and 7-fold cross-validation, where the model consistently achieved stable accuracy and precision, confirming (*C* = 1) as appropriate. The confusion matrices in Fig. 11 for both the training and test data measure classification performance and visually summarize counts for actual versus predicted values. The training precision and accuracy are about 0.8, meaning the model correctly recognizes 8 out of 10 labeled 1s (Infinite failure) and accurately detects eight labeled 1s (Infinite failure) and two labeled 0s (No failure) out of a total of 10 data points. In the testing data set, the precision is 1, and the accuracy is 0.85. Higher precision in testing data is due to the small amount of data. The testing-to-training ratio has been set to 0.2-0.5 to evaluate the precision and accuracy of the training and test data. The accuracy and precision have not varied much, which shows the SVC model is performing well.

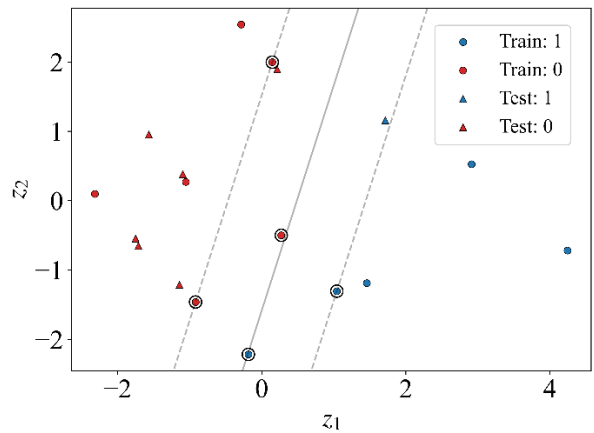

Fig.10. Binary classification of the testing and training data with SVC.

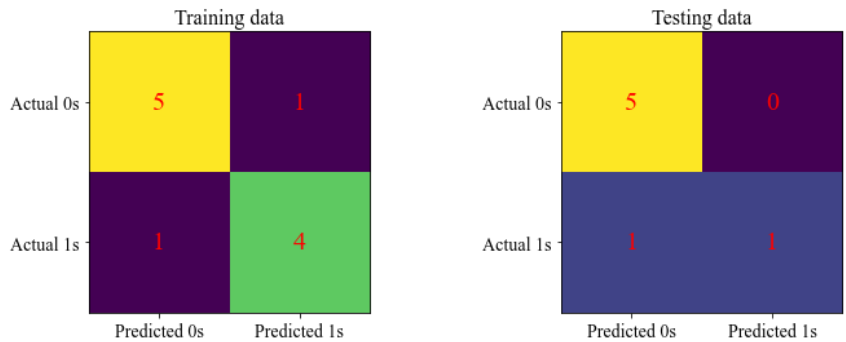

Fig.11. SVC Model performance demonstrated with confusion matrices for training and testing data.

Finally, K-Fold cross-validation is performed to validate the SVC classification. The goal is to determine how well the statistical learning procedure can be expected to perform on independent data. In this case, 7-fold cross-validation is performed on the data and run 5 times, with the data shuffled for each run (Fig. 12). The accuracy of the SVC model with $C=1$ ranges approximately from 0.75 to 0.8, indicating that the model is reliable and performs well.

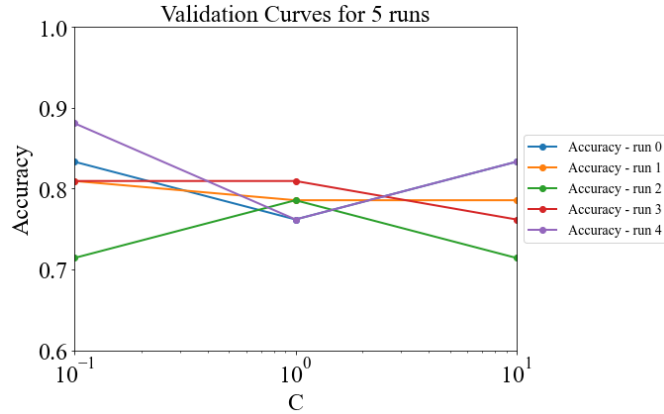

Fig.12. 7-fold cross-validation with 5 runs for SVC model

### 3.4. Logistic regression

In this research, we included the water entry value ($\Psi_{wev}$), median particle diameter size ($D_{50}$), rain intensity (RI), and gradient of the slope ($\delta$) as our independent variables. Logistic regression (LR) is applied to estimate the probability of the binary response based on the $D_{50}$, $\Psi_{wev}$, $\delta$, and $RI$ variables. The goal is to find the optimal coefficients ($\lambda$) that will maximize the likelihood function (Eq.7). In order to find the optimal coefficients, the batch gradient descent algorithm (Eq.8) is used, and the optimal coefficients are then applied to the likelihood function $\log L(\lambda)$ (Eq.7). This is an iterative method that terminates that includes an iteration index ($\upsilon$) and a learning rate ($\eta$):

$$L(\theta) = \prod_{i=1}^{m} \sigma(x)^{y^{(i)}} (1 - \sigma(x))^{1-y^{(i)}} \tag{7}$$

$$\lambda^{v+1} = \lambda^v - \eta \frac{\partial \log L(\lambda^v)}{\partial \lambda} \tag{8}$$

The four variables are combined in Eq.9 such that in the absence of rainfall ($RI=0$), the probability of the infinite failure occurrence is close to zero:

$$x = \lambda_0 + \lambda_1 D_{50}.RI + \lambda_2 \Psi_{wev}.RI + \lambda_3 \delta.RI \tag{9}$$

$\lambda_1, \lambda_2, \lambda_3$ are the optimal coefficients that are extracted using the gradient batch algorithm that is defined earlier in Eq.8 and $\lambda_0$ is the intercept of the model. The input variables are standardized to reduce prediction variance. The optimal coefficients are listed in Table 3.

Table.3 Summary of variables, coefficients and statistical evaluation parameters for testing and training data of the logistic regression (LR) model

| Features | Coefficients | Logistic regression |
|---|---|---|
|  | $\lambda_0$ | -2.38 |
| $X_1$ ($D_{50}.RI$) | $\lambda_1$ | -2.39 |
| $X_2(\Psi_{wev}.RI)$ | $\lambda_2$ | 0.53 |
| $X_3(\delta.RI)$ | $\lambda_3$ | 4.13 |
| Evaluation | Training Precision: 0. 80<br>Training Accuracy: 0.9<br>Training $TS$: 0.66 | Testing Precision: 1.00<br>Testing Accuracy: 0.9<br>Testing $TS$: 0.50 |

Fig. 13 (a-b) shows the training and testing data, as well as the probability and prediction by the LR model. The prediction is a binary classification of 0s for cases where failure did not occur ("No failure") and 1s for cases where infinite failure occurred ("*Infinite failure*"). The value of P=0.5 is set as the threshold of the probability of the occurrence of infinite failure. Suppose the predicted probability is lower than 0.5. In that case, the prediction will be class 0 ("No failure"), and if the predicted probability is 0.5 or higher, the prediction will be class 1 ("Infinite failure"). Fig. 14 shows the confusion matrix for the LR model's performance. The training precision and accuracy are about 0.8 and 0.9, and the threat score (*TS*) is 0.66. The testing precision is 1, the threat score (*TS*) is 0.5, and the accuracy is 0.9. Overall, the model performs well, and it can detect the failed cases from all predicted failed cases with high precision. In addition, the model accurately detects both failed and non-failed cases across the entire dataset.

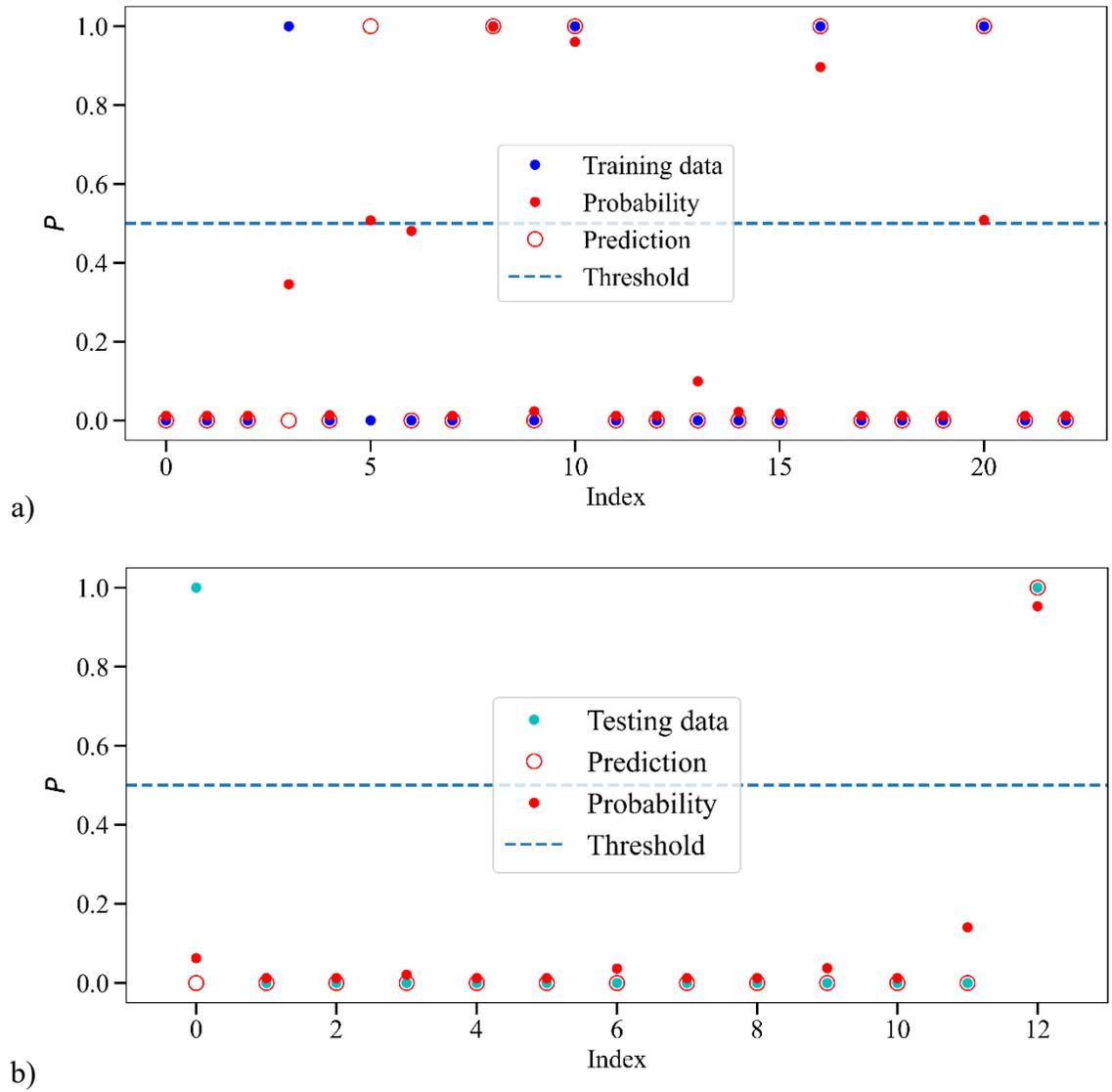

Fig.13. (a) Probability (*P*) (red filled symbols) and predictions (hollow red circles) of LR model

for (a) training and (b) testing data. The threshold that separates the two classes P= 0.5.

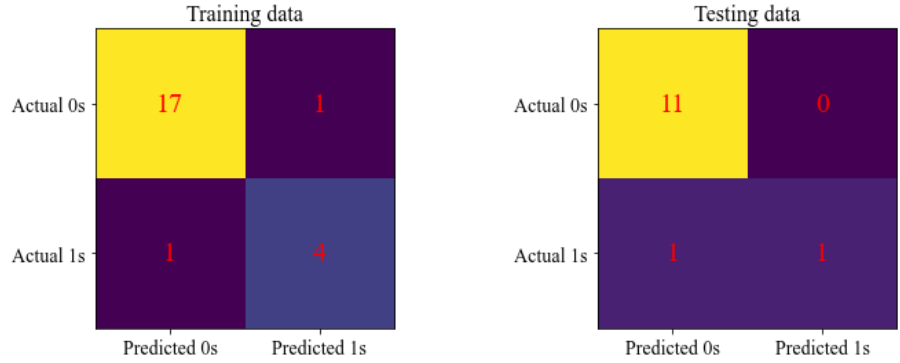

Fig..14 LR Model performance demonstrated with confusion matrices for training and testing data

### 3.5. Sensitivity analysis

Sensitivity analysis assesses the model's response to a representative range of values for a single variable while the other variables remain constant. The standard deviation method of Friedman and Santi (Staley et al., 2017; Friedman & Santi, 2014) is used to perform the analysis. In this method, the mean value of the three variables is assigned to three of the four variables, and the fourth variable is varied over a representative range. The probability analysis is then performed. This approach allows us to observe the model's sensitivity to a single variable across its range. The process has been repeated for each variable and each testing condition. The x-axis of the plots shows the standard deviation, and the y-axis shows the probability of infinite failure. Four variables are shown with blue ($D_{50}$), red ($\Psi_{\text{wev}}$), green ($\delta$), and purple ($RI$) colors.

Comparing all tests reveals that the effect of rain intensity (RI) diminishes, and the model becomes less sensitive to *RI* in tests with coarser-grain sand. For instance, by comparing Fig. 15 (a) - 17 (a), the probability of failure decreases from 1 to almost zero when the *RI* value is in the higher bound of its range. However, the model becomes more sensitive to variation in rain range in 30° slope than 20° slope (see Fig. 16 (a) and (b)), and this effect is eminently observed for medium and coarse sands by comparing columns (a) and (b) of Fig. 16 and Fig. 17. It is observed that with the same $RI$ data range, the probability of failure for medium soil reduces approximately 50% (from 1 to 0.5) (Fig. 16 (a,c,e) and (b,d,f)) and, reduces 65% (from 0.65 to 0)(Fig. 17 (a,c,e) and (b,d,f)) in coarse sand when $RI$=120 mm/hr. The model remains very sensitive to $RI$ variation in fine sand for both 20° and 30° (Fig. 15 (a-f)). Overall, fine sand shows high sensitivity to $RI$ under different conditions, and coarse sand shows the lowest sensitivity to $RI$.

Models with lower rain intensity show the second-highest sensitivity to variation in slope gradient ($\delta$) (Fig.s 15-17 (a,b)), while increasing the rain intensity increases the sensitivity of the model to variation in slope gradient ($\delta$) for all soil types (fine, medium, and coarse). Figs 15-17 (e) show that at the highest rain intensity, the model shows the greatest sensitivity to variations in $\delta$ and $D_{50}$. This suggests that $D_{50}$ and $\delta$ play a key role in controlling failure probability during high rain intensity conditions. Changes in the water entry value ($\Psi_{wev}$) have a comparatively smaller impact on the model's sensitivity, likely because the $\Psi_{wev}$ is restricted to positive values due to the hydrophobic nature of the samples (hydrophobic samples have a positive water entry value). The median grain size ($D_{50}$) has no effect on the model at low rain intensities for all sand types, as shown in Fig. 15-17 (a,b).

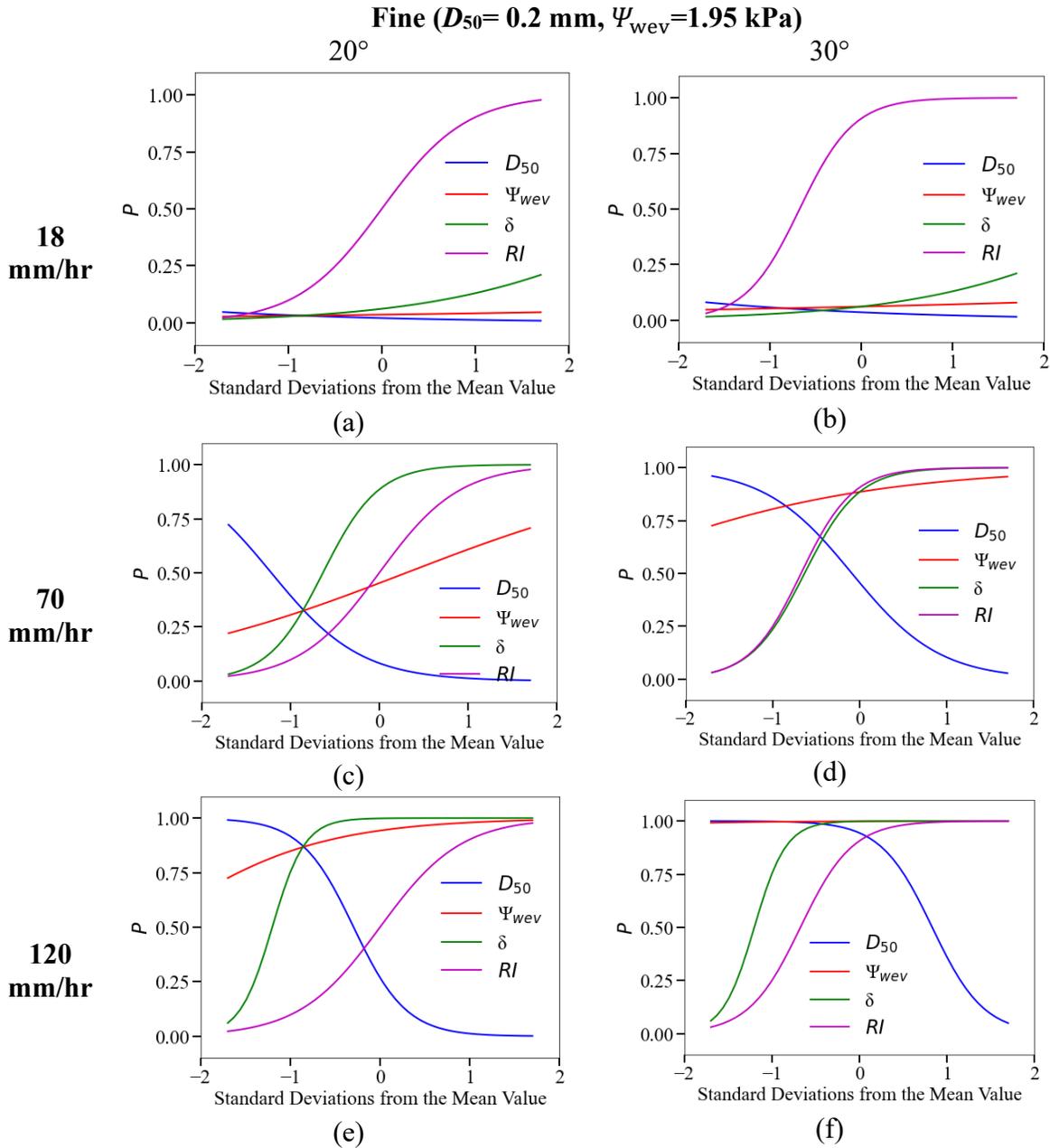

Fig.15 Sensitivity analysis results of the tests with fine sand for four features ($D_{50}$, $\Psi_{wev}$, $\delta$, RI)

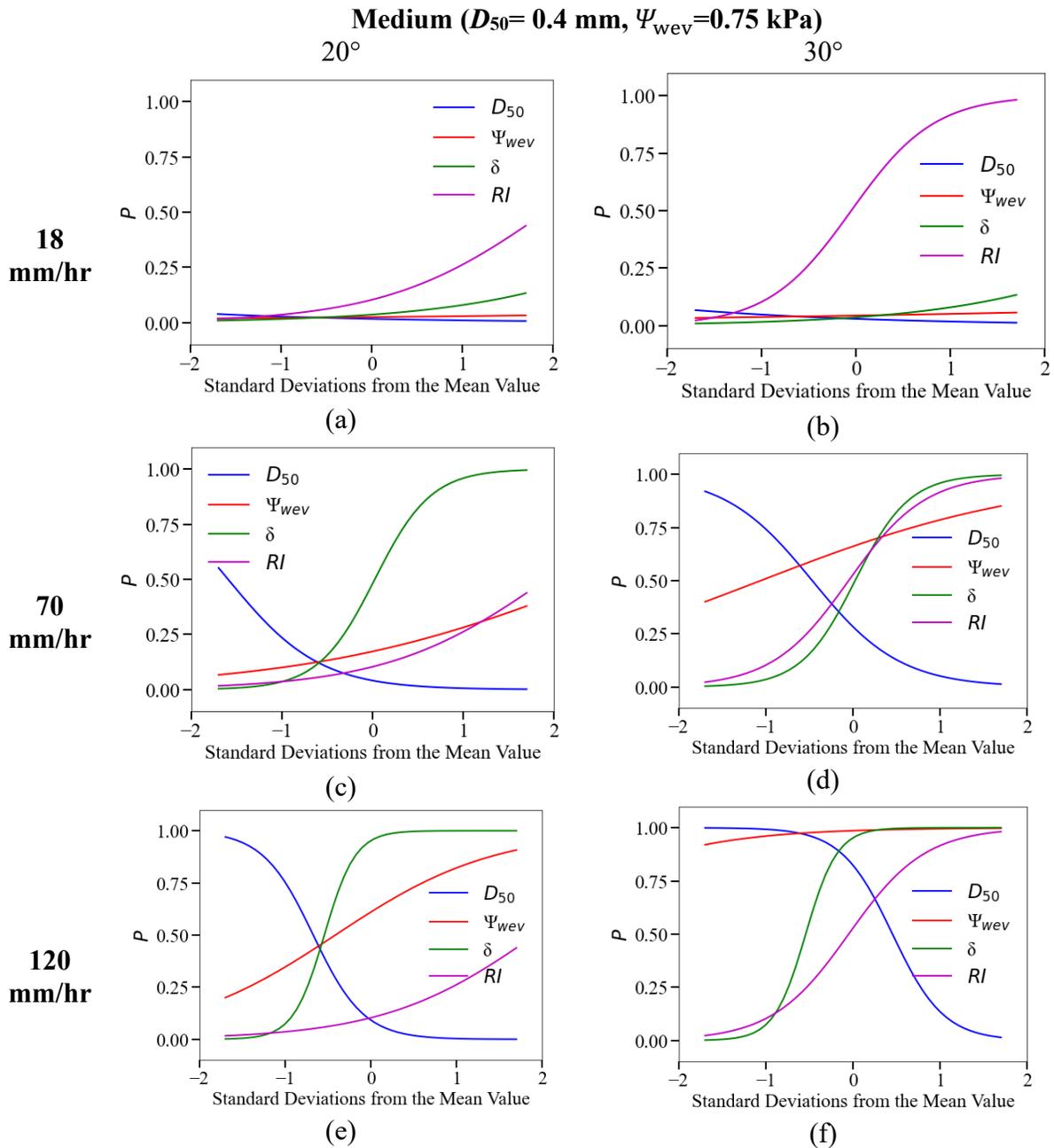

Fig.16 Sensitivity analysis results of the tests with medium sand for four features ($D_{50}$, $\Psi_{wev}$, $\delta$, RI)

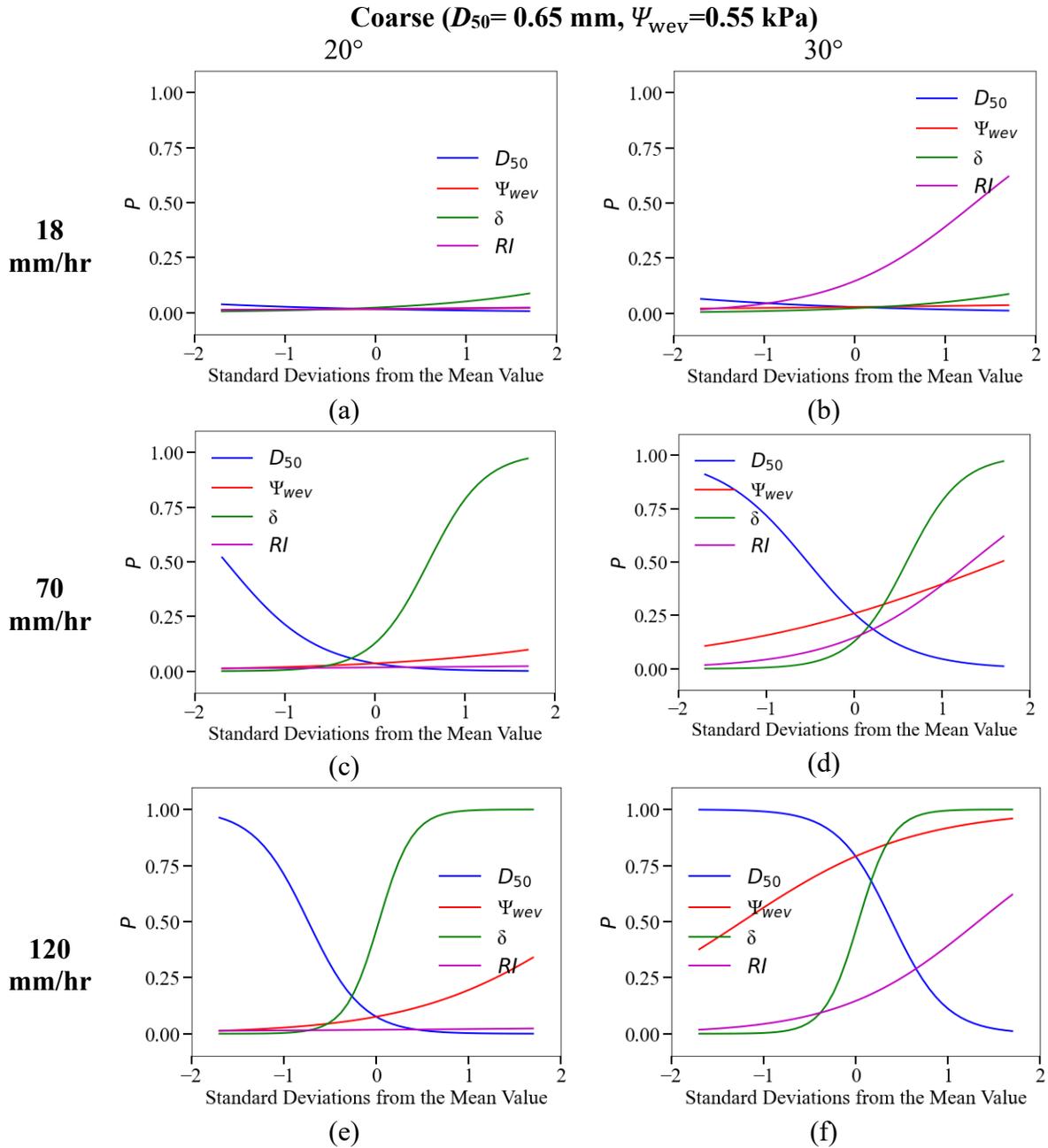

Fig.17. Sensitivity analysis results of the tests with coarse sand for four features ($D_{50}$, $\Psi_{wev}$, $\delta$, RI).

## 4. Conclusions

This study was motivated by the need to predict better the complex behavior of hydrophobic soils on burned hillslope scars after wildfires. Understanding the critical threshold values of debris flow initiation triggers can greatly improve debris flow risk management. Hydrophobic soils alter rain infiltration and runoff processes, which, in turn, affect erosion and slope stability on burned slopes. By focusing on controlled experimental data and leveraging various machine learning algorithms and methods, we aimed to predict the likelihood of failure events under varying rainfall, slope, and soil property conditions and to classify the experimental results. To achieve this, the study utilized a combination of regression, classification, clustering, and sensitivity analysis methods to model and interpret the experimental results.

For the experimental results of H-Top layout, a multiple linear regression (MLR) model was developed to predict the total erosion and total discharge of the tests under varying rain intensities ($RI$), slope gradients ($\delta$), water entry values ($\Psi_{wev}$), and median grain sizes ($D_{50}$). The experimental results from the H-Sub layout were analyzed for binary classification using logistic regression (LR), a support vector classifier (SVC), and the K-means clustering method to classify outcomes into two groups: "Infinite failure" and "No failure". A sensitivity analysis was applied exclusively to the H-Sub layout to evaluate the relative influence of the input variables. Sensitivity to RI decreases when sand grains are coarser on less inclined slopes. Specifically, the probability of failure for medium sand reduces by 50% and 65% even at very high RI=120mm/hr, while the model remains very sensitive to RI variation for fine sands. D50 and $\delta$ play a key role in controlling failure probability during high rain intensity conditions.

Although the dataset is small and may risk overfitting, the ML framework—comprising classification, clustering, and regression techniques—effectively synthesizes and visualizes the experimental data, highlighting critical variables and conditions that drive debris flow initiation. The multiple linear regression (MLR) model for the H-Top layout achieved strong agreement between predicted and measured total discharge but performed less well for total erosion, due to the low erosion observed in coarse-grained sands. This discrepancy may be attributed to the very low erosion observed in coarse-grained sands and the significant differences in erosion rates between coarse and finer sands.

The logistic regression (LR) model gives promising predictions despite a small dataset. The sensitivity analysis is performed for each test, providing a general overview of the results and facilitating more efficient interpretation. This analysis provided a clear interpretation of each variable's direct effect, which is often difficult to discern from the raw experimental data. For example, the logistic regression model showed diminishing sensitivity to RI in coarser grains; conversely, increasing RI increased its sensitivity to grain size and slope gradient.

Overall, this study identifies fine sand as the most problematic and risky sand type, compared to medium and coarse sand, for mudflow and debris flow initiation in post-wildfire areas, supporting the experimental study's conclusions. The fine sand slope, regardless of the spatial variability of the hydrophobic layer, remains the most critical for erosion and failure when coupled with a wide range of expected rainfall intensities and different slope inclinations. In addition, rainfall duration, coupled with fine sands, can be more critical during low-intensity rainfall than during high-intensity rainfall. During high rainfall intensity, the first 10 minutes of rainfall are the most critical for discharge, and most damage occurs during this initial period. In contrast, the discharged water maintains a relatively consistent percentage of the total rainfall during low-intensity rainfall. All the observations lead to the conclusion that areas with a dominant fine sand could show drastic changes in behavior after a fire, and more precautions should be considered in emergency response and evacuation plans in these areas.

**Acknowledgements**

Financial support from the Hellman Fellowship Foundation and the Regents of the University of California, San Diego (UCSD) is greatly appreciated.

**Autor contributions**

• Conceptualization: Mahta Movasat, Ingrid Tomac; Methodology: Mahta Movasat; Formal analysis and investigation: Mahta Movasat; Writing - original draft preparation: Mahta Movasat; Writing - review and editing: Mahta Movasat, Ingrid Tomac; Funding acquisition: Ingrid Tomac; Resources: Ingrid Tomac; Supervision: Ingrid Tomac.

**Funding**

Financial support from the Hellman Fellowship Foundation and the Regents of the University of California, San Diego (UCSD).

**Data availability**

Data will be available upon reasonable request.

**Declarations**

**Competing Interests**

The authors declare that they have no known competing financial interests or personal relationships that could have appeared to influence the work reported in this paper.

Goh, A. T. C., Zhang, W., Zhang, Y., Xiao, Y., & Xiang, Y. (2018). Determination of earth pressure balance tunnel-related maximum surface settlement: A multivariate adaptive regression splines approach. *Bulletin of Engineering Geology and the Environment, 77*(2), 489–500.

James, G., Witten, D., Hastie, T., & Tibshirani, R. (2013). *An introduction to statistical learning.* Springer.

Kean, J. W., Staley, D. M., & Cannon, S. H. (2011). In situ measurements of post-fire debris flows in southern California: Comparisons of the timing and magnitude of 24 debris-flow events with rainfall and soil moisture conditions. *Journal of Geophysical Research: Earth Surface, 116*(4), 1–21.

Kern, A. N., Addison, P., Oommen, T., Salazar, S. E., & Coffman, R. A. (2017). Machine learning based predictive modeling of debris in the Intermountain Western United States. *Mathematical Geosciences, 49*(6), 717–735.

Leelamanie, D. A. L., Karube, J., & Yoshida, A. (2008). Characterizing water repellency indices: Contact angle and water drop penetration time of hydrophobized sand. *Soil Science and Plant Nutrition, 54*(2), 179–187.

Martin, D. A., & Moody, J. A. (2001). Comparison of soil infiltration rates in burned and unburned mountainous watersheds. *Hydrological Processes, 15*(15), 2893–2903.

Movasat, M. (2022). *Micro to Macro Investigation of Post-Wildfire Mudflow Initiation Mechanisms*. University of California, San Diego.

Movasat, M., & Tomac, I. (2024). Kinematics of Post-Wildfire Debris Flow Initiation Mechanism: Impact of Hydrophobic Layer Spatial Variability and Particle Size. *Canadian Geotechnical Journal*, 62, 1-18.

Di Napoli, M., Marsiglia, P., Di Martire, D., Ramondini, M., Ullo, S. L., & Calcaterra, D. (2020). Landslide susceptibility assessment of wildfire burnt areas through earth-observation techniques and a machine learning-based approach. *Remote Sensing*, *12*(15), 2505.

Nikolopoulos, E. I., Destro, E., Bhuiyan, M. A. E., Borga, M., & Anagnostou, E. N. (2018). Evaluation of predictive models for post-fire debris flow occurrence in the western United States. *Natural Hazards and Earth System Sciences*, *18*(9), 2331-2343.

Nyman, P., Sheridan, G. J., Smith, H. G., & Lane, P. N. J. (2011). Evidence of debris flow occurrence after wildfire in upland catchments of south-east Australia. *Geomorphology, 125*(3), 383–401.

Pa.rsons A, Robichaud P.R., Lewis S.A., Napper C., Clark J.T. (2010). Field Guide for Mapping Post-Fire Soil Burn Severity. Technical Report RMRS-GTR-243. US Department of Agriculture, Forest Service, Rocky Mountain Research Station, Fort Collins, CO.

Puri, N., & Jain, A. (2015, March). Correlation between California bearing ratio and index properties of silt and clay of low compressibility. In *Proc. Fifth Indian Young Geotechnical Engineers Conference, Vadodara*.

Puri, N., Prasad, D. H., & Jain, A. (2018). Prediction of geotechnical parameters using machine learning techniques. *Procedia Computer Science, 125*, 509–517.